\documentclass[conference]{IEEEtran}
\usepackage{times}

\usepackage[numbers]{natbib}
\usepackage{multicol}
\usepackage[bookmarks=true]{hyperref}
\usepackage{graphicx}
\usepackage{amsmath}
\usepackage{algorithm}
\usepackage{algpseudocode}
\usepackage[margin=1in]{geometry}
\usepackage{caption}
\usepackage{subcaption}
\usepackage{amsfonts}
\usepackage{booktabs}
\usepackage{fancyhdr}
\usepackage{multirow}

\begin{document}

% paper title
\title{Can Pretrained Vision-Language Embeddings Alone Guide Robot Navigation?}

\author{
\authorblockN{Nitesh Subedi*, Adam Haroon*, Shreyan Ganguly, Samuel T.K. Tetteh, \\ Prajwal Koirala, Cody Fleming, Soumik Sarkar}
\authorblockA{
Department of Mechanical Engineering\\
Iowa State University\\
Ames, Iowa 50011, USA\\
* Equal Contribution}}

% \author{Paper ID: 30}

\maketitle

\begin{abstract}
Foundation models have revolutionized robotics by providing rich semantic representations without task-specific training. While many approaches integrate pretrained vision-language models (VLMs) with specialized navigation architectures, the fundamental question remains: can these pretrained embeddings alone successfully guide navigation without additional fine-tuning or specialized modules? We present a minimalist framework that decouples this question by training a behavior cloning policy directly on frozen vision-language embeddings from demonstrations collected by a privileged expert. Our approach achieves a 74\% success rate in navigation to language-specified targets, compared to 100\% for the state-aware expert, though requiring 3.2 times more steps on average. This performance gap reveals that pretrained embeddings effectively support basic language grounding but struggle with long-horizon planning and spatial reasoning. By providing this empirical baseline, we highlight both the capabilities and limitations of using foundation models as drop-in representations for embodied tasks, offering critical insights for robotics researchers facing practical design tradeoffs between system complexity and performance in resource-constrained scenarios. Our code is available at \href{https://github.com/oadamharoon/text2nav}{https://github.com/oadamharoon/text2nav}
\end{abstract}


\IEEEpeerreviewmaketitle

\thispagestyle{fancy}
\fancyfoot[C]{\footnotesize \textit{Accepted to Robotics: Science and Systems (RSS) 2025 Workshop on Robot Planning in the Era of Foundation Models (FM4RoboPlan)}}

\section{Introduction}
The ability to follow natural language instructions is crucial for robots to operate effectively in human environments. Instructions like "go to the red ball" simultaneously specify both a goal object and the desired action, creating a flexible interface between humans and robots~\cite{morad2024languageconditionedofflinerlmultirobot}. However, bridging the gap between language understanding and physical action presents significant challenges, as the rich semantic knowledge embedded in natural language far exceeds what robots can learn from limited task-specific experience~\cite{stone2023openworldobjectmanipulationusing}.

Traditional approaches to language-guided robot navigation typically require extensive task-specific training data and specialized architectures to ground language in visual observations. These approaches often suffer from limited generalization to novel commands or environments, as they can only recognize objects and follow instructions within predefined categories~\cite{radford2021learningtransferablevisualmodels}. The emergence of large-scale vision-language models (VLMs) like CLIP~\cite{radford2021learningtransferablevisualmodels} and SigLIP~\cite{zhai2023sigmoidlosslanguageimage} has opened new possibilities for robotics by providing rich semantic representations trained on diverse internet-scale data.

Recent work has demonstrated the benefits of integrating these foundation models into robotic systems. For example, approaches like RT-2~\cite{zitkovich2023rt} and Robotic-CLIP~\cite{nguyen2024roboticclipfinetuningclipaction} fine-tune VLMs on action data to directly generate robot commands from visual and language inputs. Other systems like CLIP-Nav~\cite{dorbala2022clipnavusingclipzeroshot} and VLM-Social-Nav~\cite{song2024vlmsocialnavsociallyawarerobot} combine VLMs with specialized navigation architectures such as mapping modules or spatial awareness components. While these approaches show impressive capabilities, they often involve complex architectures, fine-tuning procedures, or additional perception modules tailored specifically for navigation.

% \begin{figure}
%     \centering
%     \includegraphics[width=1\linewidth]{flowchart.png}
%     \caption{Enter Caption}
%     \label{fig:enter-label}
% \end{figure}

\begin{figure}[t]
    \centering
    \includegraphics[width=\linewidth]{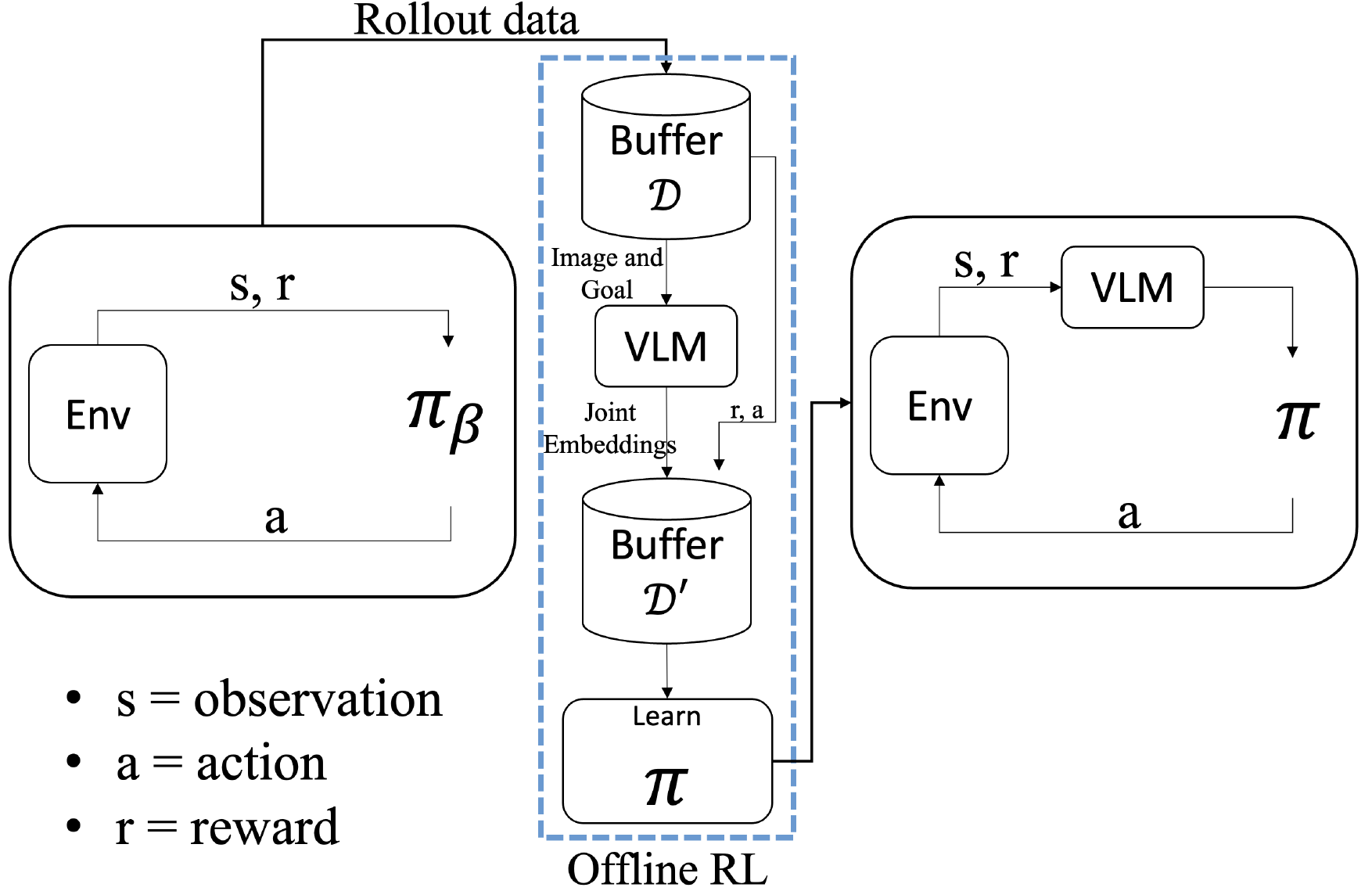}
    \caption{\textbf{Training and deployment pipeline.} A privileged expert policy $\pi_\beta$ collects demonstrations in simulation using full state observations. These are converted into joint vision-language embeddings using a pretrained VLM, forming the dataset $\mathcal{D}'$. A policy $\pi$ is then trained via behavioral cloning. At deployment, $\pi$ receives only partial observations (camera image and instruction), which are encoded by the same VLM to select actions.}
    \label{fig:system}
\end{figure}

This complexity raises a fundamental question for robotics researchers: \textit{are the rich semantic representations from pretrained VLMs already sufficient for basic navigation tasks, or are specialized architectures and fine-tuning essential?} Answering this question is critical for several reasons:

\begin{itemize}
    \item \textbf{Resource efficiency:} Complex architectures and fine-tuning require substantial computational resources that may be unnecessary if simpler approaches can achieve acceptable performance
    \item \textbf{Transfer learning:} Understanding the intrinsic capabilities of frozen embeddings provides insights into their transferability across different embodied tasks
    \item \textbf{System design:} Knowing the limitations of pretrained embeddings can guide more informed decisions about when to use specialized modules versus leveraging general-purpose representations
\end{itemize}

To address this question, we propose a minimalist approach that decouples the perception component (a frozen VLM) from the policy learning process. Rather than designing a complex end-to-end architecture, we investigate whether a simple behavior cloning policy trained on VLM embeddings can successfully navigate to language-specified targets. Our framework consists of two phases (Fig.~\ref{fig:system}): (1) training a privileged expert with full state access, and (2) distilling this expert's knowledge into a policy that operates solely on frozen vision-language embeddings.

This minimalist approach differs fundamentally from previous work in three key technical aspects. First, unlike RT-2~\cite{zitkovich2023rt} or Robotic-CLIP~\cite{nguyen2024roboticclipfinetuningclipaction}, we do not fine-tune the VLM on task-specific data, avoiding the computational cost and data requirements of adaptation. Second, unlike SpatialVLM~\cite{chen2024spatialvlmendowingvisionlanguagemodels}, we do not augment the model with specialized spatial reasoning capabilities. Third, in contrast to approaches like VLMaps~\cite{huang2023visuallanguagemapsrobot} or ZSON~\cite{majumdar2022zson}, we intentionally omit mapping or exploration modules that would provide spatial memory. Instead, we treat the VLM strictly as a frozen feature extractor, learning a reactive policy that must implicitly encode all navigation-relevant information from single-frame embeddings.

Our empirical evaluation reveals both the promise and limitations of this approach. The policy achieves a 74\% success rate in navigating to language-specified targets, demonstrating that pretrained embeddings can indeed support basic grounding of language to visual targets. However, the policy takes significantly longer paths (3.2× on average) compared to the expert, indicating limitations in spatial reasoning, planning, and memory. These findings provide valuable insights for researchers seeking to balance simplicity and performance in language-guided robot navigation systems.

Our contributions include:
\begin{itemize}
    \item A minimalist framework for language-guided navigation that uses frozen vision-language embeddings as the sole representation for policy learning
    \item An empirical evaluation demonstrating that pretrained embeddings alone can achieve a 74\% success rate in semantic target navigation
    \item Analysis of the performance gap between our VLM-guided policy and the privileged expert, revealing the strengths and limitations of pretrained embeddings for navigation
    \item Insights into future directions for effectively leveraging foundation models in embodied robotics
\end{itemize}

\section{Methodology}
Our approach consists of two phases: (1) an expert demonstration phase using privileged information, and (2) a learning phase where a policy is trained on the expert data using vision-language representations. After training, the learned policy is deployed using only its onboard sensors and the language instruction.

\subsection{Phase 1: Privileged Expert Policy}
\begin{figure}[t]
    \centering
    \includegraphics[width=\linewidth]{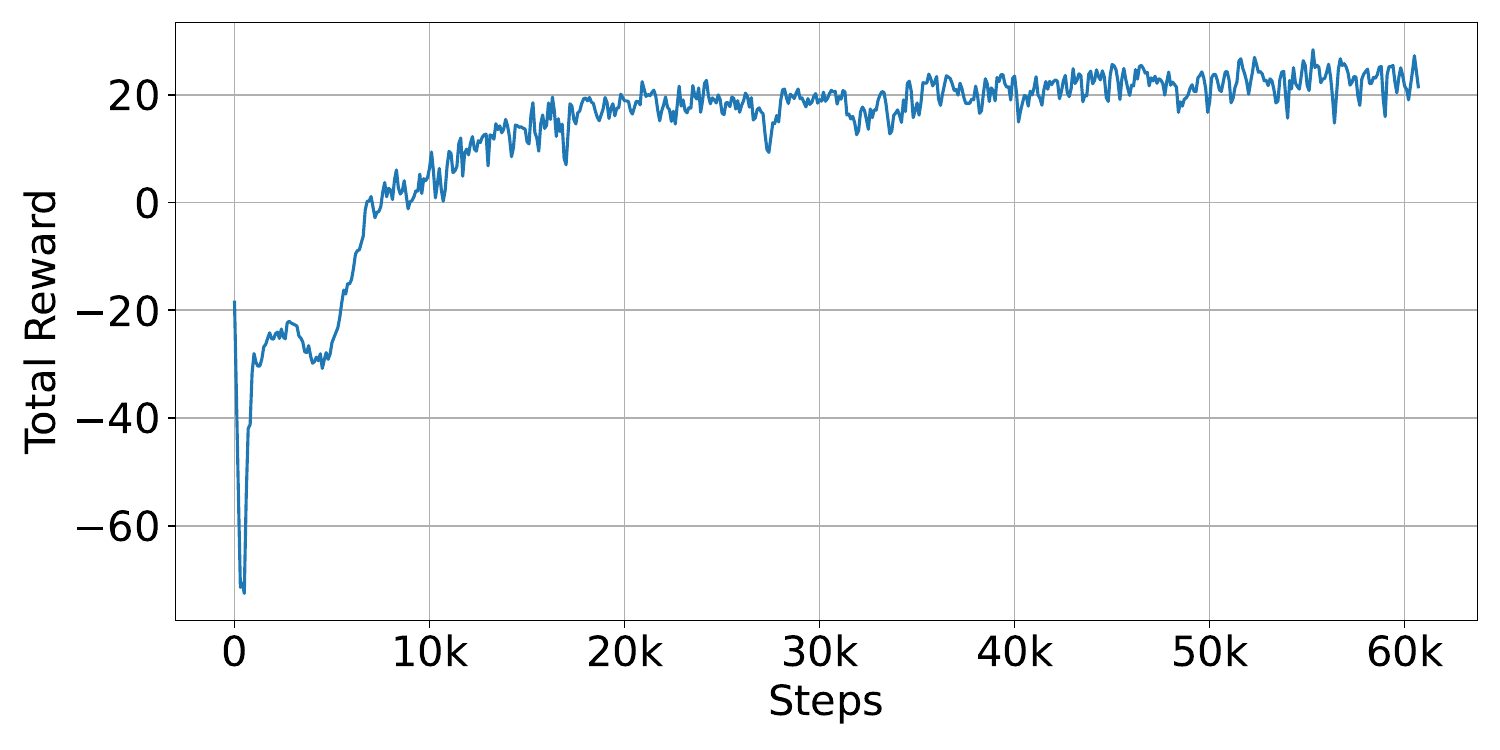}
    \caption{\textbf{Expert policy training reward.} Reward plot for the expert policy $\pi_\beta$, which quickly learns to successfully navigate to specified targets using privileged state information.}
    \label{fig:reward_plot_pi_beta}
\end{figure}

In Phase 1, we develop an expert behavioral policy $\pi_\beta(s, g)$ that can reliably navigate to a goal, where $s$ denotes the full state and $g$ denotes the goal specification. We represent $g$ as a one-hot indicator of the target sphere's identity (among five possibilities) along with its position in the environment. The expert knows both which object is the target and its precise location, making navigation a straightforward control problem. We implement $\pi_\beta$ using Proximal Policy Optimization (PPO)~\cite{schulman2017proximalpolicyoptimizationalgorithms}.

The expert policy effectively functions as a perfect GPS-based navigator. In our preliminary evaluations, it achieves a 100\% success rate, consistently reaching the correct sphere and stopping within 0.1m of the target. Importantly, $\pi_\beta$ receives the target identity directly rather than processing language instructions—this is acceptable since its role is solely to generate demonstrations, not to be deployed at test time.

Using $\pi_\beta$, we collect a dataset of navigation trajectories. Each trajectory begins with the robot at a random location and orientation, with one of the five colored spheres randomly chosen as the goal. We generate natural language commands that include both semantic color identification and spatial positioning relative to the robot's current orientation (e.g., "The target is the red ball which is to your left. Move toward the ball." or "The target is the blue ball which is straight ahead. Move toward the ball.") for each target. These relative spatial cues are determined based on the angular position of the target relative to the robot's heading. The expert controls the robot until it reaches the goal, and we record:
\begin{itemize}
    \item Camera images from the robot's perspective
    \item Language instructions for the target
    \item Expert actions (wheel velocity commands)
    \item Reward signals based on distance to goal
\end{itemize}

The dataset $\mathcal{D}$ consists of approximately 500 trajectories (100 per target object), averaging 5-8 seconds of navigation each, resulting in tens of thousands of state-action pairs for training.

\begin{algorithm}[t]
\caption{Training Language-Conditioned Navigation Policy via Behavioral Cloning}
\label{alg:bc_pipeline}
\begin{algorithmic}[1]
\Require Simulator $\mathcal{S}$, expert policy $\pi_\beta$, pretrained VLM (SigLIP), number of episodes $N$
\Ensure Learned policy $\pi$ that maps VLM joint embeddings to robot actions

\State Initialize dataset $\mathcal{D} \gets \emptyset$
\For{$i = 1$ to $N$}
    \State Sample initial state $s_0$ in $\mathcal{S}$
    \State Sample goal object $g$
    \State Determine spatial instruction $l$ from $g$ and robot pose
    \State Initialize trajectory buffer $\tau \gets \emptyset$
    \While{episode not terminated}
        \State Get expert action $a_t = \pi_\beta(s_t, g)$
        \State Capture RGB image $I_t$
        \State Store $(I_t, l, a_t)$ in $\tau$
        \State Apply $a_t$ in $\mathcal{S}$ to get $s_{t+1}$
    \EndWhile
    \State Add $\tau$ to dataset $\mathcal{D}$
\EndFor

\State Initialize BC dataset $\mathcal{D}' \gets \emptyset$
\ForAll{$(I_t, l, a_t) \in \mathcal{D}$}
    \State Image embedding $\mathbf{v}_t = \text{L2Norm}(\text{VLM}_{\text{img}}(I_t))$
    \State Text embedding $\mathbf{u}_t = \text{L2Norm}(\text{VLM}_{\text{text}}(l))$
    \State Joint embedding: $\mathbf{o}_t = \text{L2Norm}(\mathbf{v}_t + \mathbf{u}_t)$
    \State Add $(\mathbf{o}_t, a_t)$ to $\mathcal{D}'$
\EndFor

\State Train policy $\pi$ to minimize loss:
\[
    \mathcal{L}(\theta) = \mathbb{E}_{(\mathbf{o}, a) \sim \mathcal{D}'}\left[\| \pi_\theta(\mathbf{o}) - a \|^2 \right]
\]
\State \Return trained policy $\pi$
\end{algorithmic}
\end{algorithm}

\subsection{Vision-Language Model Selection}
\label{subsec:vlm_selection}
To guide our choice of vision-language model, we first assessed whether pretrained VLMs encode not only semantic object information but also spatial relationships relevant to navigation tasks. We conducted an empirical comparison of three prominent models: ViLT (VQA-finetuned)~\cite{kim2021viltvisionandlanguagetransformerconvolution}, CLIP (ViT-B/32)~\cite{radford2021learningtransferablevisualmodels}, and SigLIP (So400M-Patch14-384)~\cite{alabdulmohsin2024gettingvitshapescaling}.

We designed a spatial distance test based on joint embeddings of images and language prompts. For each reference image containing a colored sphere, we generated a natural language instruction. Each 256×256 image was divided into an even 3×3 spatial grid, creating nine distinct regions for localizing objects. For each reference image, we selected two additional images containing the same colored object—one where the object appeared in the same spatial grid cell (same spatial semantics), and one where it appeared in a different cell (different spatial semantics). We controlled for background and lighting conditions to isolate the effect of spatial positioning.

We computed the cosine distance between the joint image-text embeddings of the reference image and each comparison image across x different configurations. Lower cosine distance indicates greater semantic and spatial similarity. If a vision-language model encodes spatial semantics, we expect lower distances when spatial positioning is consistent, even when the same object and instruction are present in both images.

As shown in Table~\ref{tab:spatial_vlm_eval}, all three models exhibited sensitivity to spatial positioning, with lower cosine distances for image pairs sharing the same spatial cell. Notably, SigLIP demonstrated the most pronounced distinction between same and different spatial contexts (26.5\% difference), suggesting stronger spatial sensitivity in its joint embedding space compared to CLIP (13.0\% difference) and ViLT (16.3\% difference).

\begin{table}[hbt!]
\centering
\resizebox{\linewidth}{!}{%
\begin{tabular}{lccc}
\toprule
\textbf{Model} & \textbf{Spatial Relation} & \textbf{Mean Distance ($\mu$)} & \textbf{Std Dev ($\sigma$)} \\
\midrule
\multirow{2}{*}{ViLT (VQA)} 
& Same Cell & 0.0380 & 0.0374 \\
& Different Cell & 0.0442 & 0.0447 \\
\midrule
\multirow{2}{*}{CLIP (ViT-B/32)} 
& Same Cell & 0.0276 & 0.0130 \\
& Different Cell & 0.0312 & 0.0144 \\
\midrule
\multirow{2}{*}{SigLIP} 
& Same Cell & 0.0245 & 0.0118 \\
& Different Cell & 0.0310 & 0.0145 \\
\bottomrule
\end{tabular}
}
\caption{\textbf{VLM spatial sensitivity analysis.}  Cosine distance between joint image-text embeddings across spatial configurations. Lower values indicate greater embedding similarity given a task. All models show smaller distances for same spatial positions, but SigLIP shows the most pronounced separation.}
\label{tab:spatial_vlm_eval}
\end{table}

While this test does not fully assess complex spatial reasoning capabilities required for navigation planning, it confirms that pretrained VLMs inherently encode some degree of spatial information in their joint embeddings without explicit spatial training. This finding, combined with SigLIP's efficient architecture and strong performance on downstream tasks with fewer parameters~\cite{alabdulmohsin2024gettingvitshapescaling}, led us to select it as our vision-language backbone.

\subsection{Vision-Language Embedding with SigLIP}
Based on our comparative analysis, we leverage SigLIP~\cite{zhai2023sigmoidlosslanguageimage} to bridge the gap between high-dimensional sensory inputs and the learning algorithm. SigLIP consists of an image encoder and a text encoder that project images and text into a shared 1152-dimensional latent space where semantically related concepts have high cosine similarity. The model's sigmoid-based contrastive loss function offers improved stability compared to CLIP's~\cite{radford2021learningtransferablevisualmodels} softmax-based approach, making it particularly suitable for robotic applications.

For each timestep in our dataset, we compute a joint embedding that fuses visual and language information:

\begin{enumerate}
    \item The image $I$ and instruction $L$ are processed through SigLIP's pretrained encoders
    \item The resulting features are L2-normalized
    \item The normalized features are summed and normalized again to produce a joint embedding vector $\mathbf{o}$
\end{enumerate}

This vector $\mathbf{o}$ serves as a compact representation of both what the robot sees, what it has been instructed to do, and the spatial relationship between the robot and target. The inclusion of relative spatial descriptors in the language instructions enables the joint embedding to capture both semantic object identity and directional guidance. Critically, the SigLIP model remains completely frozen throughout our pipeline—we make no updates to its weights during training.

\subsection{Behavioral Cloning Policy Learning}
With the processed dataset $\mathcal{D}' = \{(\mathbf{o}_t, a_t)\}$, we train a policy $\pi(a | \mathbf{o})$ via behavioral cloning~\cite{liang2024constrainedbehaviorcloningrobotic}. The policy is implemented as a feedforward neural network that takes the joint embedding vector $\mathbf{o}$ as input and outputs a two-dimensional action $a = (\omega_1, \omega_2)$ representing normalized wheel velocity commands.

We train the policy to minimize the mean squared error between its predictions and the expert's actions:
\[
\mathcal{L}(\theta) = \mathbb{E}_{(\mathbf{o},a) \sim \mathcal{D}'}\left[\|\pi_\theta(\mathbf{o}) - a\|^2\right]
\]

We optimize this loss using Adam with an initial learning rate of $1 \times 10^{-3}$, decaying over time. The network converges quickly due to the large quantity of near-optimal demonstrations.

An important consideration is that the expert's actions are conditioned on privileged information that the student policy lacks. The student policy must learn behaviors like "turn towards the red object when seeing it" and "explore when the target is not visible" purely from the demonstration data, without ever receiving explicit state information. At deployment, the policy takes as input the current camera image and language instruction, processes them through SigLIP to obtain the joint embedding $\mathbf{o}$, and outputs action commands in a closed-loop manner.

\section{Experimental Setup}
\begin{figure}[t]
    \centering
    \includegraphics[width=\linewidth]{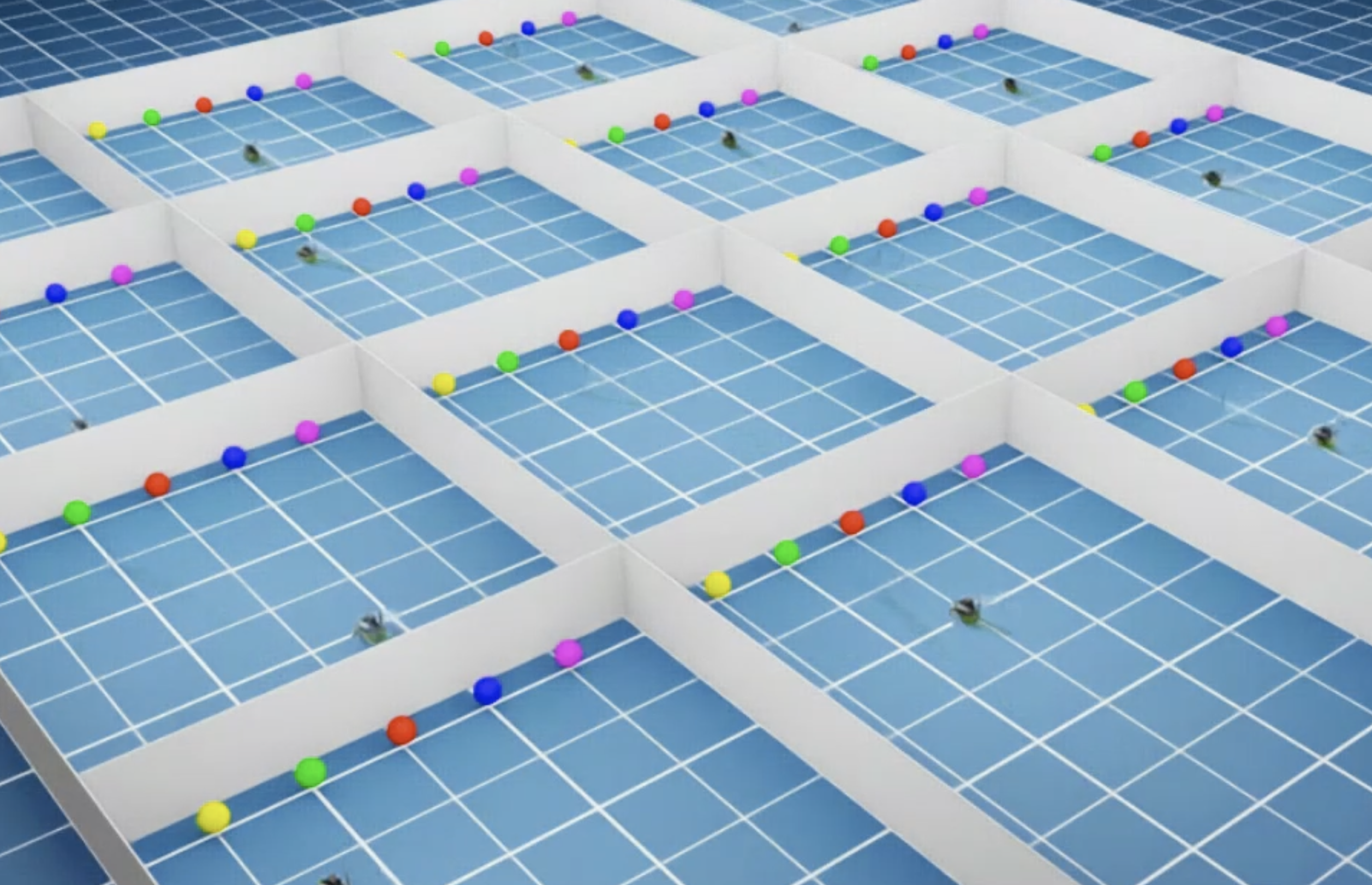}
    \caption{\textbf{Simulation environment.} Simulation environment showing the Jetbot with multiple colored spheres. During evaluation, the robot must navigate to a specific colored sphere based on language instructions that include both semantic color identification and relative spatial cues using only RGB camera input and pretrained vision-language embeddings.}
    \label{fig:sim_env}
\end{figure}

\paragraph{Simulation Environment} 
We conduct all experiments in NVIDIA Isaac Sim~\cite{isaacsim} and NVIDIA Isaac Lab~\cite{mittal2023orbit}, a high-fidelity robotics simulation framework. Our code is available at \href{https://github.com/oadamharoon/text2nav}{https://github.com/oadamharoon/text2nav}. The environment consists of a flat 3m × 3m arena with five colored spheres (red, green, blue, yellow, and pink) of 10cm diameter positioned at fixed locations for each episode. The robot starts at a random location and orientation. No obstacles are present, focusing the challenge on identifying the correct target and navigating efficiently toward it among visually similar distractors.

\paragraph{Robot Platform} 
The agent is modeled after the NVIDIA Jetbot, a small differential-drive robot with a forward-facing RGB camera. The camera provides a first-person view at 256×256 resolution. The robot's action space consists of two-dimensional angular velocity commands that directly control the wheels. Each action is applied for a short timestep (1/60 seconds) before the next observation is captured.

\paragraph{Task Definition} 
At the start of each episode, one of the five colored balls is randomly chosen as the target. A natural language instruction is generated that includes both the target color and its relative spatial position based on the angular offset between the robot's heading and target location: "The target is the [color] ball which is to your [left/right/straight ahead]. Move toward the ball." The robot must navigate to the correct target using only its egocentric RGB images and the language instruction. Success is defined as reaching within a small distance (0.1m) of the correct target within a maximum of 1000 timesteps.

\paragraph{Evaluation Protocol}
We evaluate both the expert policy $\pi_\beta$ and the learned policy $\pi$ on 100 test episodes with randomized starting positions. For each episode, we record:
\begin{itemize}
    \item Success rate (reaching the correct target)
    \item Number of timesteps to completion
    \item Cumulative reward
    \item Trajectory efficiency (compared to optimal paths)
\end{itemize}

\section{Results}
Our evaluation reveals a substantial performance gap between the expert policy with privileged state access and the vision-language embedding-based policy (Table~\ref{tab:results}). The expert policy $\pi_\beta$ achieves a perfect 100\% success rate, while the learned policy $\pi$ succeeds in 74\% of test episodes.
\begin{table}[h]
    \centering
    \begin{tabular}{lrr}
    \toprule
    Metric &  $\pi_\beta$ &  $\pi$ \\
    \midrule
    Success Rate (\%)   & 100 & 74.0 \\
    Avg Timesteps  & 113.97 & 369.4 \\
    Max Timesteps  & 154.00 & 828.00 \\
    Min Timesteps  & 73.00 & 216.00 \\
    \bottomrule
    \end{tabular}
    \caption{\textbf{Performance comparison between policies.} The learned policy $\pi$ using only vision-language embeddings achieves 74\% success but requires significantly more timesteps than the privileged expert $\pi_\beta$, even in successful cases. This highlights both the potential and limitations of using pretrained embeddings alone for navigation.}
    \label{tab:results}
\end{table}

\paragraph{Success Rate Analysis} 
The 74\% success rate of the vision-language policy is notable, as it demonstrates that pretrained embeddings alone can guide a robot to the correct target in a majority of cases without any architecture specifically designed for navigation or spatial reasoning. The policy correctly grounds language instructions in visual observations and generates appropriate actions to reach the target—all through the lens of a frozen vision-language model and a simple feedforward network.

\paragraph{Efficiency Gap} 
Despite its reasonable success rate, the learned policy exhibits significantly lower efficiency compared to the expert. When successful, $\pi$ takes on average 369.4 timesteps to reach the goal—3.2× more than the expert's average of 114.0 timesteps. Even the best-case performance of $\pi$ (216 timesteps) is approximately triple the expert's minimum (73 timesteps), while the worst case requires 828 timesteps compared to the expert's maximum of 154.

\begin{figure}[t]
    \centering
    \includegraphics[width=\linewidth]{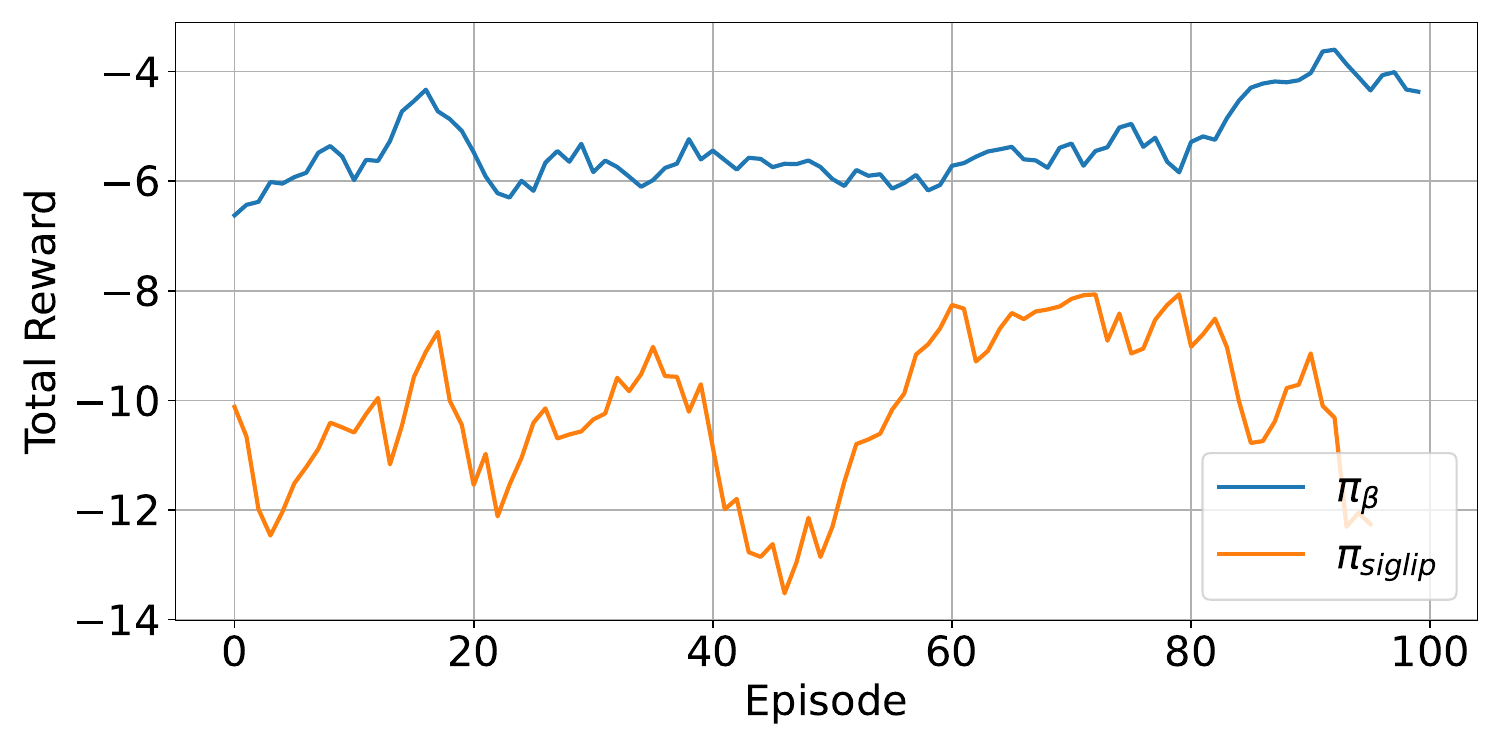}
    \caption{\textbf{Reward comparison.} Cumulative rewards comparison between the expert ($\pi_\beta$, blue) and VLM-based policy ($\pi$, orange). The learned policy shows greater variability and consistently lower rewards due to inefficient navigation and occasional failures.}
    \label{fig:reward_plots}
\end{figure}

\paragraph{Reward and Consistency}
The reward comparison (Fig.~\ref{fig:reward_plots}) further illustrates the performance gap. The learned policy's reward curve shows considerable episode-to-episode variation, with rewards ranging from approximately -8 to -13, indicating inconsistent performance across different starting configurations. In contrast, the expert maintains stable, high rewards across all test episodes. This variability suggests that the VLM-based policy struggles more with certain spatial arrangements or viewing angles than others.

\paragraph{Learning Dynamics}
The behavioral cloning loss for SigLIP (Fig.~\ref{fig:bc_loss}) shows rapid initial convergence followed by more gradual improvement, eventually stabilizing around 0.06. This suggests that while the policy learns to approximate the expert's actions in many situations, there remains a persistent gap—likely corresponding to scenarios where the expert relied on privileged information not captured in the visual-linguistic embeddings.

\subsection{Baseline Evaluation: Multi-Model Comparison}

While our main results demonstrate SigLIP's effectiveness, we conducted a controlled comparison to validate our VLM selection methodology and understand the impact of different vision-language architectures, using identical training procedures and evaluation protocols across all three candidate models. We trained separate behavioral cloning policies using embeddings from ViLT (VQA-finetuned), CLIP (ViT-B/32), and SigLIP, evaluating each on 100 episodes for fair statistical comparison. 

\begin{figure}[t]
    \centering
    \includegraphics[width=1.0\linewidth]{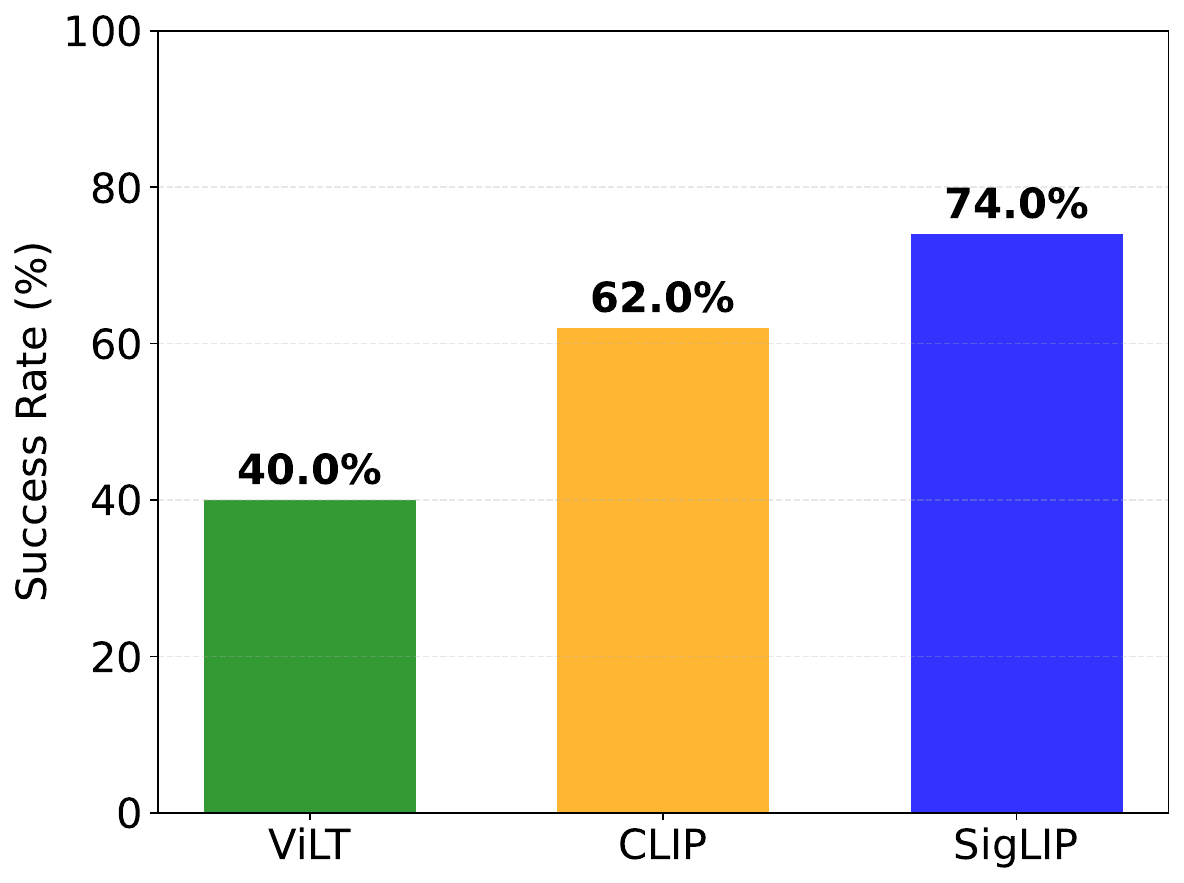}
    \caption{\textbf{Multi-model performance comparison.} Success rates across three vision-language models using identical BC training procedures (n=100 episodes each). SigLIP demonstrates superior performance, achieving 74.0\% success rate compared to CLIP's 62.0\% and ViLT's 40.0\%.}
    \label{fig:model_comparison}
\end{figure}

The results reveal a clear performance hierarchy: SigLIP (74.0\% success) $>$ CLIP (62.0\%) $>$ ViLT (40.0\%), as shown in Figure~\ref{fig:model_comparison}. Remarkably, SigLIP achieved both the highest success rate and superior efficiency, requiring an average of only 369.4 timesteps for successful episodes compared to CLIP's 417.6 and ViLT's 472.0 timesteps. This dual superiority in both success and efficiency strongly validates our methodological choice of SigLIP as the vision-language backbone.

\paragraph{Embedding Dimensionality and Performance}
The performance ranking aligns closely with embedding dimensionality: SigLIP's 1152-dimensional representations significantly outperform ViLT's 768-dimensional and CLIP's 512-dimensional embeddings. This suggests that richer representational capacity translates directly to better navigation performance, even when using frozen embeddings without task-specific adaptation.

\begin{figure}[t]
    \centering
    \includegraphics[width=\linewidth]{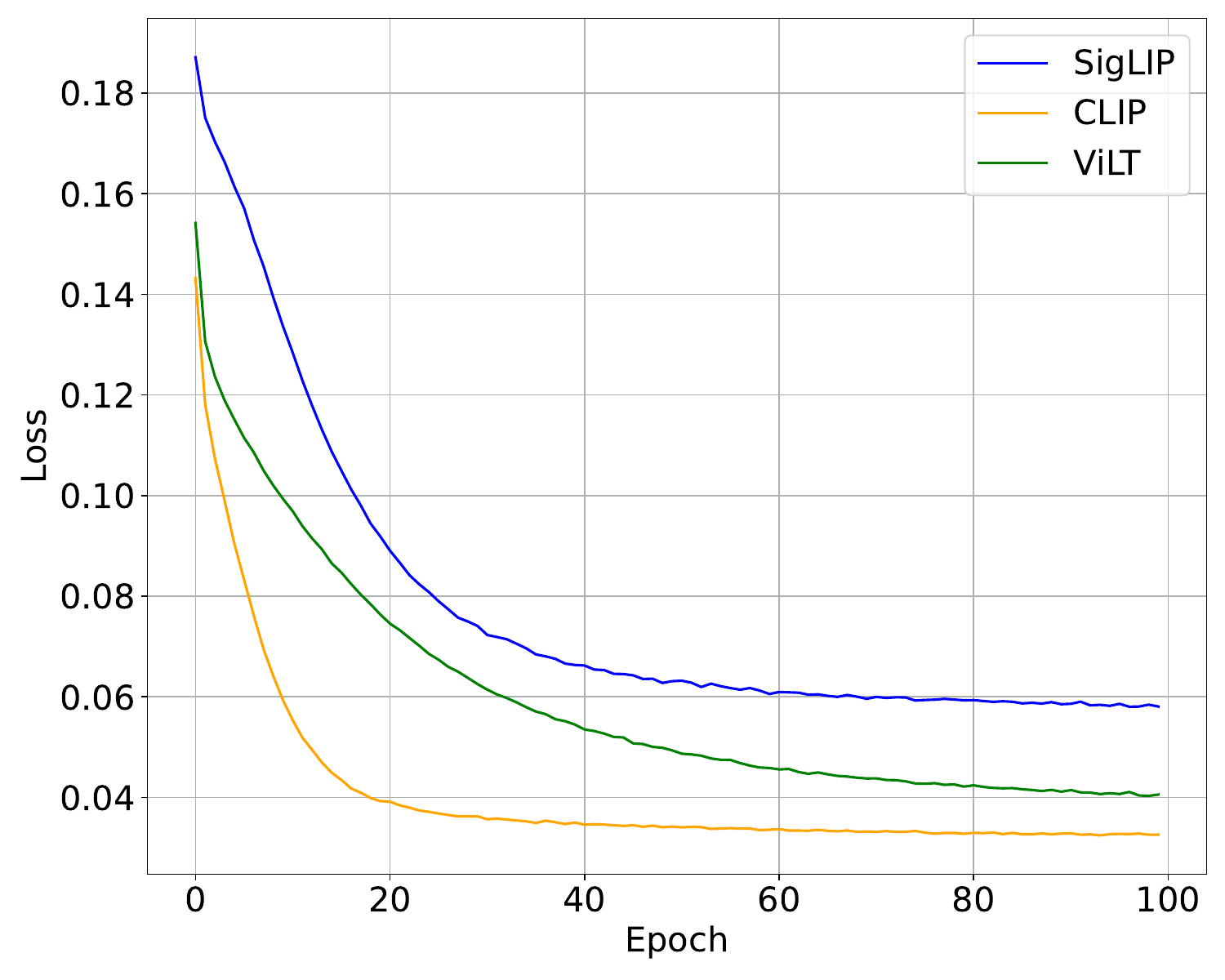}
    \caption{\textbf{Behavioral cloning loss across VLM architectures.} SigLIP achieves superior navigation performance despite higher BC loss than ViLT and CLIP, suggesting that BC loss does not predict downstream task performance for embodied applications.}
    \label{fig:bc_loss}
\end{figure}

\paragraph{BC Loss vs. Navigation Performance}
Interestingly, the navigation performance ranking inversely correlates with behavioral cloning loss (Fig.~\ref{fig:bc_loss}). ViLT and CLIP achieved lower BC training losses, while SigLIP exhibited higher BC loss but superior actual navigation performance. This confirms our earlier hypothesis that BC loss measures imitation fidelity rather than task performance—SigLIP's richer 1152-dimensional embeddings are harder for the BC network to compress into action predictions, but they contain more navigation-relevant information that enables better spatial reasoning.

\paragraph{Spatial Sensitivity Validation}
These results validate our spatial sensitivity analysis from Section~\ref{subsec:vlm_selection}, where SigLIP demonstrated the most pronounced spatial distinction (26.5\% cosine distance difference between same/different spatial positions) compared to CLIP (13.0\%) and ViLT (16.3\%). The navigation performance hierarchy directly mirrors the spatial sensitivity ranking, confirming that inherent spatial understanding in vision-language embeddings translates to practical navigation capabilities.

\paragraph{Statistical Significance}
The 95\% confidence intervals reveal statistically significant differences between models: SigLIP (74.0\% [65.4\% - 82.6\%]), CLIP (62.0\% [52.5\% - 71.5\%]), and ViLT (40.0\% [30.4\% - 49.6\%]). The non-overlapping confidence intervals between SigLIP and ViLT, and minimal overlap between SigLIP and CLIP, demonstrate that the performance differences are statistically robust.

\paragraph{VQA Fine-tuning Limitations}
ViLT's poor performance (40.0\% success), despite being fine-tuned for visual question answering, suggests that task-specific fine-tuning on non-embodied datasets may not transfer effectively to robotic navigation. The VQA fine-tuning appears to have specialized ViLT for discrete question-answering rather than the continuous spatial reasoning required for navigation, highlighting the importance of representation generality for embodied tasks.

These findings provide several key insights for using VLMs in navigation: (1) larger embedding dimensions generally improve performance when sufficient training data is available, (2) models with stronger inherent spatial sensitivity achieve better navigation results, (3) BC loss is not predictive of downstream task performance, and (4) general-purpose vision-language models may outperform task-specific fine-tuned variants for embodied applications requiring spatial reasoning.

\section{Discussion and Conclusion}

Our work addresses a fundamental question in embodied AI: can pretrained vision-language embeddings alone serve as a sufficient representation for navigation policies? The results provide a nuanced answer: while these embeddings enable basic language-guided navigation (74\% success rate), they show clear limitations in efficiency and consistency compared to state-aware approaches. This performance gap offers valuable insights into both the capabilities and limitations of using foundation models as drop-in representations for embodied tasks.

\subsection{Semantic Grounding vs. Spatial Reasoning}
Our results demonstrate that pretrained vision-language embeddings excel at semantic grounding—connecting language descriptions to visual observations. The policy successfully differentiates between colored targets based solely on the joint embeddings of images and instructions. This capability aligns with the core strength of models like SigLIP~\cite{zhai2023sigmoidlosslanguageimage} and CLIP~\cite{radford2021learningtransferablevisualmodels}, which are trained specifically to align visual and linguistic concepts.

However, the efficiency gap reveals that these embeddings struggle with more complex spatial reasoning and planning. While our VLM selection study showed that SigLIP encodes basic spatial sensitivity (distinguishing same vs. different object positions), this level of spatial understanding appears insufficient for efficient navigation. The 3.2× longer paths taken by our policy suggest that while it can identify targets, it lacks the temporal memory and higher-order spatial reasoning needed for direct navigation. Unlike specialized approaches that incorporate mapping modules~\cite{huang2023visuallanguagemapsrobot} or explicit spatial reasoning components~\cite{chen2024spatialvlmendowingvisionlanguagemodels}, our minimalist policy must infer spatial relationships purely from a sequence of independent embeddings without explicit spatial representation.

This distinction between basic spatial sensitivity and complex spatial reasoning highlights an important consideration for robotics researchers: while foundation models offer powerful semantic representations and some inherent spatial understanding out-of-the-box, they may need to be complemented with dedicated spatial reasoning mechanisms for tasks requiring efficient path planning and exploration. Recent work in latent space planning~\cite{hafner2019learninglatentdynamicsplanning, rosetebeas2022latentplanstaskagnosticoffline} offers promising directions for bridging this gap by learning structured representations that support both semantic understanding and planning.

\subsection{Prompt Engineering Sensitivity}
An important consideration in our approach is the sensitivity to prompt design. Our initial experiments with simple color-only instructions (e.g., "Go to the red ball") yielded significantly lower performance. The inclusion of relative spatial cues ("The target is red ball which is to your left. Move toward the ball.") proved crucial for enabling the VLM embeddings to ground both semantic and spatial information effectively. This highlights the importance of prompt engineering when leveraging foundation models for embodied tasks, as the linguistic framing directly impacts the model's ability to extract navigation-relevant features from the joint embedding space.

\subsection{Performance-Complexity Tradeoffs}
Our minimalist approach highlights an important tradeoff in robotics: simplicity versus performance. More complex approaches like RT-2~\cite{zitkovich2023rt} or NaVILA~\cite{cheng2025navilaleggedrobotvisionlanguageaction} can achieve higher performance through fine-tuning or specialized architectures but require substantial computational resources and engineering effort. In contrast, our approach using frozen embeddings and simple behavioral cloning demonstrates that reasonable performance can be achieved with minimal complexity.

This tradeoff is particularly relevant for resource-constrained applications where training compute or specialized hardware may be limited. Our results suggest that for basic navigation tasks with clear visual targets, pretrained embeddings alone may provide sufficient capabilities without the need for extensive fine-tuning or complex architectures. However, for tasks requiring efficient navigation or complex reasoning, the additional complexity of specialized approaches may be justified.

Recent work in student-teacher distillation~\cite{kim2025distillingrealizablestudentsunrealizable} and zero-shot policy transfer~\cite{jang2022bczzeroshottaskgeneralization} offers promising directions for better balancing this tradeoff. By more effectively distilling the privileged expert's knowledge into a realizable student policy, it may be possible to achieve performance closer to the expert while maintaining the simplicity of our approach.

\subsection{Implications for Future Research}
Our findings suggest several promising directions for future research at the intersection of foundation models and robotics:

\begin{itemize}
    \item \textbf{Hybrid architectures}: To address the repeated circling and target confusion failures we observed, combining the semantic richness of pretrained embeddings with explicit spatial memory could enable the policy to maintain consistent object identification across viewpoints. World models~\cite{hafner2024masteringdiversedomainsworld} offer a promising framework for integrating perception with spatial reasoning.
    
    % \item \textbf{Temporal integration}: To overcome the local exploration loops we observed, mechanisms like recurrent networks or attention-based architectures could help the policy aggregate information over time, reducing repeated visits to already-explored regions and building on the basic spatial sensitivity we detected in VLMs.
    
    \item \textbf{Task-specific representation learning}: The confusion between visually similar targets suggests that generic vision-language embeddings may not optimally encode the distinctions most relevant for navigation. Approaches like R3M~\cite{nair2022r3muniversalvisualrepresentation} could provide embodiment-aware representations that better capture navigation-relevant features while building on the spatial sensitivity already present in models like SigLIP.
    
    \item \textbf{Data-efficient adaptation}: The timeout failures indicate that frozen embeddings struggle with systematic exploration. Lightweight adaptation techniques could align pretrained representations with navigation-specific requirements without the full computational cost of fine-tuning.
    
    % \item \textbf{Offline reinforcement learning}: Extending our behavior cloning approach with offline RL techniques~\cite{levine2020offlinereinforcementlearningtutorial, bundele2024scalingvisionandlanguagenavigationoffline} might enable the policy to improve beyond the demonstrated trajectories and potentially close the efficiency gap.
\end{itemize}

\subsection{Conclusion}
Our work provides an important empirical baseline for understanding the capabilities and limitations of pretrained vision-language embeddings in embodied navigation tasks. By demonstrating that these embeddings alone can achieve a 74\% success rate in following language instructions, we highlight their potential as lightweight representations for robotics. However, the significant efficiency gap compared to privileged experts reveals the need for additional inductive biases or specialized components to achieve expert-level performance.

This minimalist approach provides robotics researchers with a practical baseline for evaluating when to use foundation models as-is, versus when to invest in specialized components. As foundation models continue to advance in capability and efficiency, understanding their intrinsic strengths and limitations becomes increasingly important for effective system design. By isolating and quantifying the specific contribution of vision-language embeddings to navigation performance, our work enables more informed architectural decisions at the intersection of foundation models and embodied robotics.

\section{Acknowledgment}
This work is funded from NSF-USDA COALESCE grant \#2021-67021-34418.

\bibliographystyle{plainnat}
\bibliography{references}

\end{document}